\documentclass{article}

\usepackage{graphicx} 
\usepackage{markdown}
\usepackage{biblatex}
\usepackage{hyperref}

\title{An Approach to Checking Correctness for Agentic Systems}
\author{Thomas J Sheffler}
\date{June 2025}

\begin{document}

\maketitle


\begin{abstract}

This paper presents a temporal expression language for monitoring AI agent behavior, enabling systematic error-detection of LLM-based agentic systems that exhibit variable outputs due to stochastic generation processes. Drawing from temporal logic techniques used in hardware verification, this approach monitors execution traces of agent tool calls and state transitions to detect deviations from expected behavioral patterns.
  
Current error-detection approaches rely primarily on text matching of inputs and outputs, which proves fragile due to the natural language variability inherent in LLM responses. The proposed method instead focuses on the sequence of agent actions—such as tool invocations and inter-agent communications—allowing verification of system behavior independent of specific textual outputs.

The temporal expression language provides assertions that capture correct behavioral patterns across multiple execution scenarios. These assertions serve dual purposes: validating prompt engineering and guardrail effectiveness during development, and providing regression testing when agents are updated with new LLMs or modified logic.

This approach is demonstrated using a three-agent collaborative system from Google's Agent Development Kit (ADK), where agents coordinate to solve multi-step reasoning tasks. When powered by large, capable models, all temporal assertions were satisfied across many test runs. However, when smaller models were substituted in two of the three agents, executions violated behavioral assertions, primarily due to improper tool sequencing and failed coordination handoffs. The temporal expressions successfully flagged these anomalies, demonstrating the method's effectiveness for detecting behavioral regressions in production agentic systems.

This approach provides a foundation for systematic monitoring of AI agent reliability as these systems become increasingly deployed in critical applications.

\end{abstract}

\section{Introduction}

An AI agent is a software entity that employs an LLM (Large Language Model) to perform tasks and respond to natural language interactions.  Agents specialize an LLM for a particular task with specific prompt instructions and augment the model with a conversation history, memory, and tools.  The LLM provides the engine for the logic of the agent.  Today's LLMs can plan over multiple steps how to approach a task and can make use of ``tools'' (invoked through function calls) to retrieve additional information, to store results in a memory or database, or possibly to alter the environment of the running system.

LLMs can exhibit unpredictability in their planning and computations however.  Because they produce variable outputs,  LLMs may solve similar tasks in different ways at different times.  This behavior can be surprising, but is also useful for searching a solution space.   More predictable behavior can be obtained by using prompt engineering techniques that help focus the solution space of the LLM \cite{openaiGPT41Prompting}.

Various means have been developed to verify that an agent is behaving as desired.  Guardrail code can encode specific checks on agent output based on the input text or elements of the entire conversation.  Execution traces are often captured for analysis and formal methods employing assertions over temporal expressions have also been developed.

\begin{figure}
  \includegraphics[width=\columnwidth]{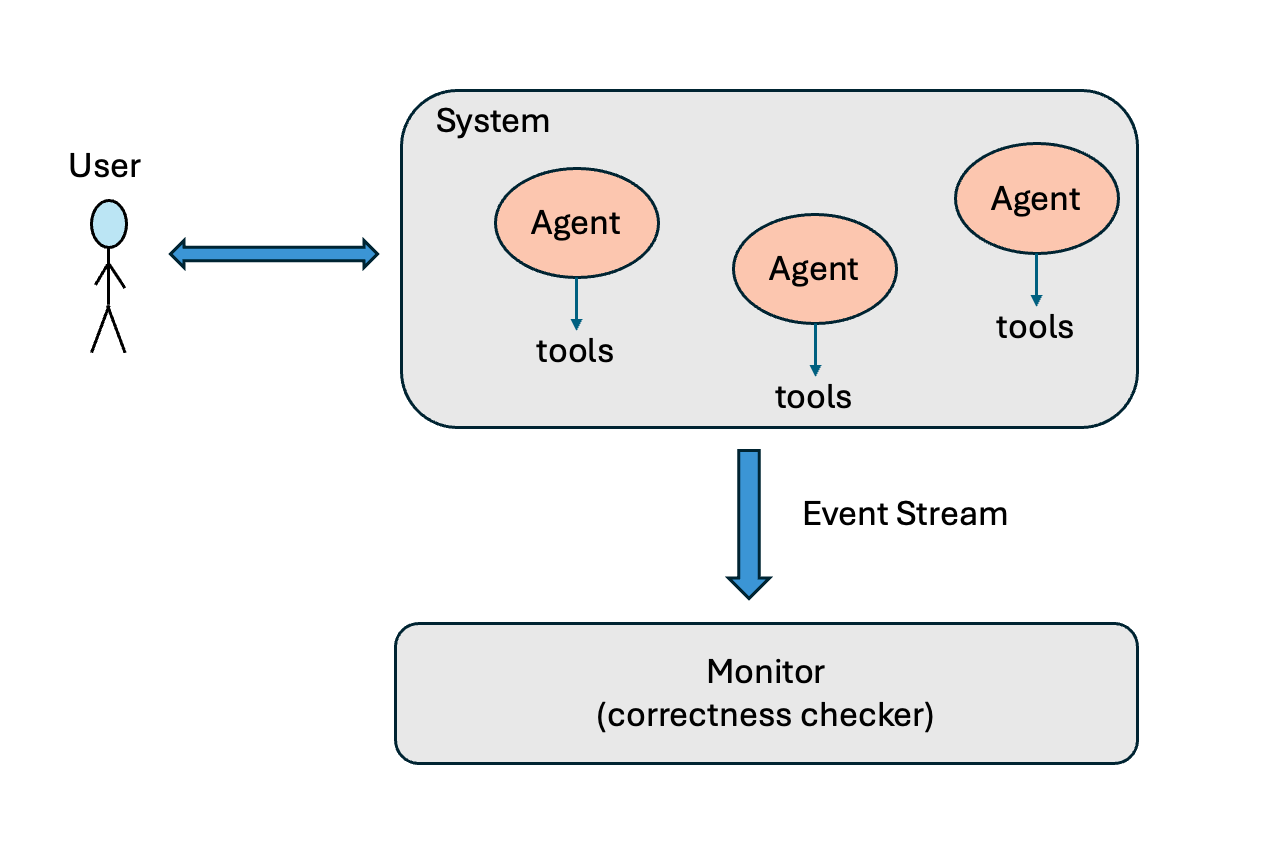}
  \caption{A Monitor runs alongside the Agentic System}
  \label{fig:monitor}
\end{figure}

This paper presents an event-driven temporal expression framework that can be executed in popular agent toolkits that are running today.  Given an agentic system, a correctness checker executes in an associated monitor, as shown in Fig.~\ref{fig:monitor}.  The monitor receives events from the running agentic system and alerts when unexpected event traces are received.

The paper demonstrates the validity of the approach by showing the use of the framework in a  multi-agent system.  Incorrect event sequences are detected when the LLM executes a disallowed transition.  The temporal expression language is succinct and applicable to a wide variety of circumstances.  This paper suggests that assertions of the temporal properties of agentic systems should be developed early in the design process as an agent is created and that such assertions should remain part of the execution environment of an agent throughout its lifetime.

\subsection{Temporal Logic and Assertions in Event-driven systems}

Linear Temporal Logic (LTL) is a mathematical language for describing linear-time properties\cite{Baier2008}.  It has been used in theorem proving (show that a property is satisfied for a given system model) and model checking (check that all possible executions of a system model satisfy a formal specification).

An LTL formula may consist of atomic propositions, the logical operators NOT, AND and OR, and the temporal operators $X$ (neXt) and $\cup$ (Until).  LTL formulae over the set $AP$ of atomic propositions are formed according to the following grammar:

$$
   \phi ::= true  \mid  a   \mid   \phi1 \land \phi2  \mid \neg \phi   \mid  X \phi   \mid   \phi_1 \cup \phi_2
$$  
where $a \in AP$.

\begin{figure}
  \includegraphics[width=\columnwidth]{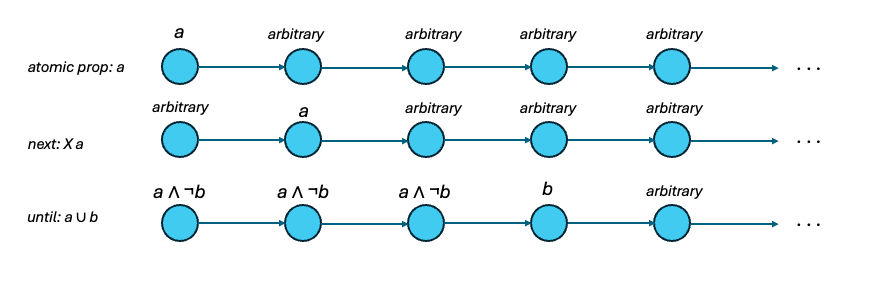}
  \caption{Meaning of LTL Temporal Operators}
  \label{fig:ltl}
\end{figure}

A intuitive visual explanation of these operators is shown in Fig.~\ref{fig:ltl}.  The assertion of an atomic property 'a' at the current time means it is currently satisfied.  The proposition 'X a' means that the atomic property is satisfied in the next time step.  The proposition 'a U b' means that a is satisfied in the steps between now and when 'b' is satisfied.

LTL has been used as formalism to specify real-time properties in synchronous systems, such as those defined with a system clock.  In this type of system the "next" operator X, can be defined to have a "timed" interpretation.  "X $\phi$" states that at the next time step, $\phi$ holds.

For instance, the formula
$$
  X X X \phi
$$
means that $\phi$ holds after (exactly) 3 time steps.

Temporal Expression Languages with their roots in the formalism of LTL have been developed for use with logic simulation and verification.  These include the Property Specification Language (PSL) \cite{IEEE_1850_2010,doulosKnowHowPsl,wikipediaPropertySpecification} and Temporal Expressions of the Specman/e Verification Language \cite{IEEEStd1647, asicworldEventsTemporal, wikipediaverificationLanguage}.  Each of these approaches adds syntactic sugar beyond LTL to ease the expression of properties in synchronous systems.  A semi-formal approach was also developed in the package called SystemVerilog Assertions (SVA) \cite{systemVerilogAssertionsHandbook,doulosSystemVerilogAssertions,sutherlandSystemVerilogAssertions}.  When used in a simulation context, temporal expression languages such as these execute assertions alongside a DUT (Design under Test) to continually monitor a path expression against the sequence of states the design produces.

Temporal expression assertion checkers have shown to be powerful tools for representing the desired behavior of systems under development by reducing errors in testbench code.  Without a temporal expression assertion language, a verification team would need to code elements such as counters and state machines to check properties.  This code is often delicate and prone to errors.  Instead, temporal assertion languages provide powerful and compact means expressing complex sequences over time.

\subsection{Related Work on Agents and Temporal Assertions}


Formal systems for verifying properties of AI agentic systems have been explored in various contexts. Bosse et al. \cite{conf/atal/BosseLB06} described a method for capturing state traces of agent behavior and analyzing those traces using software tools. Their formulation of dynamic properties incorporates Linear Temporal Logic, with implementation running offline using traces of events from the agentic system.

Model checking is a technique by which formal properties of a system are verified against a specification.  In \cite{10.1109/RO-MAN46459.2019.8956429} the authors describe the MCMAS model checker to verify logical formulae against the decision-making behavior of an agent.  Their system can produce a counter-example of formulae that are not satisfied.  Model checking approaches typically run offline, against a formal representation of the system, rather than concurrently with a running implementation.

The "Verified AI" project from Berkeley \cite{berkeleylearnverifyVerified} describes "Verified Artificial Intelligence (AI) as the goal of designing AI-based systems that have strong, ideally provable, assurances of correctness with respect to mathematically-specified requirements."   One of the tools produced from this group is the "VerifAI" Python package.  This package implements a temporal logic called "Metric Temporal Logic".  This package has many uses, one of which is temporal-logic falsification (bug-finding).  That particular use case has many similarities with what is presented here.  The implementation approach is different, and lacks the trace information shown here.  The software package requires many dependencies.

Another Python project implementing Metric Temporal Logic is \cite{githubGitHubDoganuluspythonmonitors}.  This package provides a means to monitor formal specifications over temporal sequences.  The stated goal of that package is similar to the one presented in this paper.  Counterexample traces produced are different from those here, and are not as extensive.

\subsection{Organization of this Paper}

The next section gives a brief overview of the the Oroboro temporal expression package.  It describes how predicates map an existing Python system into the temporal framework, and how system events are used to advance the stepping of time.

Section 3 describes an experiment embedding temporal expressions in a three-agent demonstration system distributed as part of the Google Agent Development Kit (ADK) \cite{googleAgentDevelopmentKitStep3}.  The characteristics of the example system are presented and the mechanism for developing the temporal expressions for this system are explained.

Section 4 discusses the experimental results of running the instrumented system.  Example execution traces from a correct system are captured and presented.  For a system with errors, the results show how an erroneous sequence of tool calls is detected and reported in real time.

Section 5 presents some concluding remarks about the test methodology and suggests directions for future development.

\section{Oroboro Temporal Expressions}

Oroboro is a Python package implementing a temporal expression language inspired by those of the property checkers of simulated Hardware Description Languages (HDLs)\cite{githubShefflerOroboro}.  It implements a useful subset of operators and provides enough syntactic sugar to ease the writing of temporal expressions for common cases.  The Oroboro temporal expression package aims to be usable by users who are not necessarily familiar with formal specifications or model checking.

Oroboro temporal expressions can observe state in a Python environment.  Expressions are built from foundational "predicates" that compute the truth of a fact in the running Python environment at the current time.  A predicate is simplest temporal expression.  More complex expressions are built from predicates and temporal expression operators.

An Oroboro temporal expression either matches a series of events starting at a given time, or it fails.  In either case, the evaluation of the expression produces a "trace" which is similar to a "capture" in a string-based regular expression package.  The trace shows the start and end time of the match, and decomposes the top-level match into the matches of its sub-expressions.  A temporal expression failure is similar, and shows where the expression failed to match.

The evaluation of an Oroboro temporal expression is driven by a sampling event that is defined by the user.  The signaling of the sampling event causes each of the predicates of the temporal expression to be evaluated and for time to move forward by one step.  A sampling event keeps a counter that is incremented each time it is signaled. The sample event count is used in traces to report when a temporal expression started and ended.  The inclusion of the event count in the trace is a usability aid for the user.

The evaluation of an Oroboro temporal expression is driven by a sampling event that is defined by the user.  The signaling of the sampling event causes each of the predicates of the temporal expression to be evaluated and for time to move forward by one step.  Each occurrence of the sampling event is numbered, and this number is reported in the traces.  The inclusion of the event count in the trace is a usability aid for the user.

\subsection{Operators}

The base operator for creating a instantaneous temporal expression is the `Pred` operator.  This operator takes a Python callable that returns True or False and produces a single time-step temporal expression that succeeds or fails.  More complex temporal expressions are built using the operators described below.

\begin{description}
\item[\texttt{Pred(fn)}]
  (Predicate) \texttt{Pred} succeeds at the current event if \texttt{fn()} returns True and fails otherwise.

\item[\texttt{a + b}]
  (Concatenation)  The expression \texttt{(a+b)} succeeds if \texttt{a} succeeds at this event, and \texttt{b} succeeds at the next event.

\item[\texttt{a >> b}]

  (Conditional)  The expression \texttt{(a >> b)} succeeds if \texttt{a} fails, or if a succeeds on this event and \texttt{b} succeeds on the next event.

\item[\texttt{a | b}]

  (Alternation) The expression \texttt{(a | b)} succeeds if either \texttt{a} succeeds on this event or \texttt{b} succeeds on this event.
  
\item[\texttt{a \& b}]

  (Conjunction)  The expression \texttt{(a \& b)} succeeds if \texttt{a} and \texttt{b} both succeed, and they succeed with the same start and end time.
  
\item[\texttt{a / b}]

  (Fusion)  The expression \texttt{(a / b)} succeeds if \texttt{a} matches, and the \texttt{b} matches in the same cycle as \texttt{a}.

\item[\texttt{a * n}, or  \texttt{a * (n, m)}]

  (Repeat)  This operator implements successive concatenation.  The expression \texttt{(a * n)} matches if a matches exactly n times in succession.  The second argument may also be a tuple, \texttt{(n,m)}.   The expression \texttt{(a * (n,m))} matches if a matches at least \texttt{n} times in succession but no more than \texttt{m} times.
\end{description}  
  
\subsection{Traces}

The evaluation of a temporal expression beginning at a particular sampling event either succeeds or fails after some finite number of occurrences of the sampling event.  In this section a temporal expression will be defined and example traces that it produces will be explained.

An example temporal expression, \texttt{te}, is defined below and is built from two predicates.  A predicate named \texttt{ok} that succeeds at every occurrence of the sampling event.  A predicate called \texttt{a} is defined relative to a stimulus bit-vector called \texttt{avals}.  It will succeed where the bit-vector has a '1' and fail where the bit-vector has a '0'. (The detailed definition of predicate \texttt{a} is not included in this paper for the sake of brevity.)

The temporal expression called \texttt{te} matches if there are two occurrences of predicate \texttt{a} in a row, or a sequence that begins with an \texttt{a} and matches \texttt{a} again two cycles later, with anything in between.

\begin{verbatim}
ok = Pred( lambda d: return True )
a = Pred( ... )
te = (a + a) | (a + ok + a)
\end{verbatim}

The input  bit-vector stimulus is shown below.  The cycle number is in a comment, and the bit-vector is defined in the \texttt{avals} vector.  This stimulus is then applied to the system.

\begin{verbatim}
    # cycle     1 2 3 4 5 6 7 8
    avals =   [ 0 1 1 1 0 1 0 0 ]
\end{verbatim}

When evaluated at cycle 4, a match is detected.  The trace associated with the match is shown below.  This match states that the top-level expression ('$\mid$' - alternation) matched starting at cycle 4 and ending at cycle 6.  Two '+' (concatenation) sub-expressions matched.  The first matched starting at cycle 4 and ending at cycle 5 with the sequence 'a, OK'.   The top-level '+' (concatenation) matched starting at cycle 4 and ending at cycle 6 by concatenating an 'a' at cycle 6 onto the sub-expression from cycle 4 to 5.

\begin{verbatim}
MATCH
| (4/6) (40/60) 0 {}
  + (4/6) (40/60) 0 {}
    + (4/5) (40/50) 0 {}
      preda (4/4) (40/40) 0 {}
      __ok (5/5) (50/50) 0 {}
    preda (6/6) (60/60) 0 {}
\end{verbatim}

When evaluated at cycle 6, a failure would be detected at cycle 8.  The trace associated with the failures is shown below.  At cycle 7 the LHS of the alternation oeprator failed when two successive \texttt{a} predicates failed.  Since the top-level operator was ('$\mid$' - alternation, both of its sub-expressions need to fail for the entire operator to fail.  This happened at cycle 8 when the RHS of the alternation operator failed when the predicate \texttt{a} did not succeed at cycle 8.

\begin{verbatim}
FAILURE
| (6/8) (60/80) 0 {}
  + (6/7) (60/70) 0 {}
    preda (6/6) (60/60) 1 {}
    preda (7/7) (70/70) 0 {}
  + (6/8) (60/80) 0 {}
    + (6/7) (60/70) 1 {}
    preda (8/8) (80/80) 0 {}
\end{verbatim}

\section{Adding Assertions to an Agentic System}

The Oroboro temporal expression framework was embedded into the Google Agent Development Kit (ADK) and used to model some of the expected execution paths in a demonstration three-agent system released by Google.

The multi-agent agent consisted of three separate agents.

\begin{itemize}
\item a main "Weather" agent respond to questions about the weather by calling a \texttt{get\_weather} tool and to delegate greetings and farewlls to sub-agents.
\item a "Greeting" agent to provide the user with a greeting by calling its \texttt{say\_hello} tool
\item a "Farewell" agent that provides a nice farewell to the user by calling its \texttt{say\_goodbye} tool
\end{itemize}

The demonstration system is part of a tutorial on using the ADK.  As a tutorial it is informative, illustrating many of the capabilities of the ADK and how they fit together.
The three-agent example lacks a strict specification though.  The intuitive behavior of each of the agents is described, but corner cases and failure handling is not addressed since that is not the focus of the tutorial.  However, an attempt to more completely define this system is possible.

\begin{figure}
  \includegraphics[width=\columnwidth]{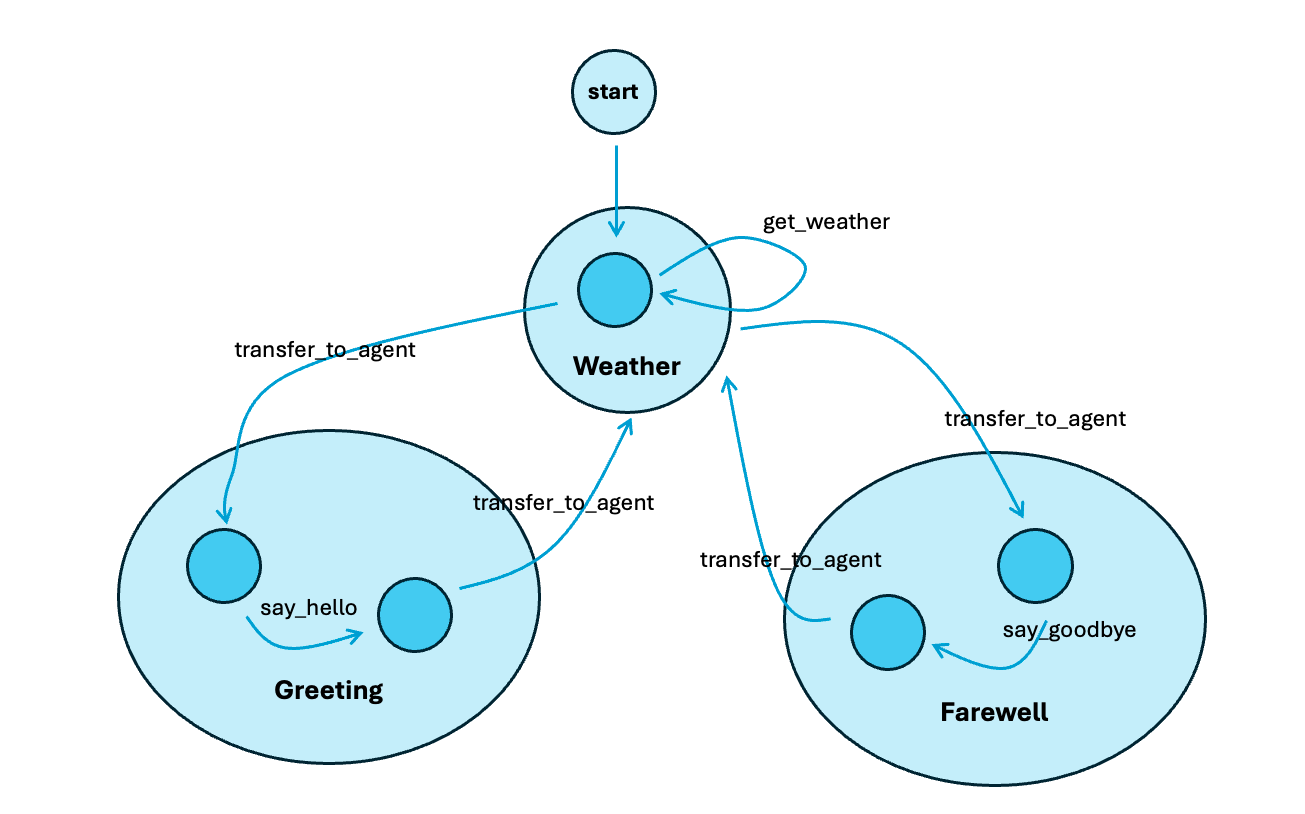}
  \caption{The State Transition Graph of the Three-Agent System}
  \label{fig:statetransitions}
\end{figure}

Fig.~\ref{fig:statetransitions} illustrates a state transition graph developed for this simple system.  It has five states.   States are shown grouped by the agent in control:  either the Weather, Greeting or Farewell agent.  If a user query about the weather is received, it can be answered by the Weather agent which can call the \texttt{get\_weather} tool.  If a greeting is desired, control must be transferred to the Greeting agent where it can call its \texttt{say\_hello} too.   Similarly, if a farewell is desired, control must be transferred to the Farewell agent so it can call its \texttt{say\_goodbye} tool.  Excursions to the Greeting and Farewell agents should immediately return to the Weather agent.

Having produced the state transition graph here, it would be possible to encode it as a state machine in the Python system executing the three-agent system and use it as an executable specification for checking correctness.  The use of such a state machine would be one approach to checking the correctness of the system.  In many cases, however, a complete state machine is an over-specification.  Temporal expressions can be used to encode portions of the state machine of the system and leave other parts unspecified, or up to the implementation.  The next section illustrates how this is done.

\subsection{Instrumenting the System}

In the three-agent system described, there are three associated tools that are to be called by each agent.  There is also another tool provided by the ADK that allows one agent to transfer control to another.  This tool is called \texttt{transfer\_to\_agent} and takes as an argument the name of the destination agent.

Hooks provided by the ADK (\texttt{before\_tool\_call} and \texttt{after\_tool\_call}) were used to to catch calls to these tools.  The following predicates were defined for use by the temporal expression assertions.  The sampling event used by the system was the call to \texttt{before\_tool\_call}.  The resulting system produces a "one-hot" encoding of predicates in which only one may be true at a time.  (This encoding technique is not a limitation of the Oroboro system or the ADK, but is simply the characteristic of this particular three-agent system.)

\begin{itemize}
\item xferToGreeting
\item xferToFarewell
\item xferToWeather
\item sayHello
\item sayGoodbye
\item getWeather
\end{itemize}

The first three predicates mark when a \texttt{transfer\_to\_agent} tool call occurs and its destination is as shown.  The last three mark when \texttt{say\_hello}, \texttt{say\_goodbye} or \texttt{get\_weather} tool calls are emitted.

With the small family of predicates defined, we could write some temporal expressions that should be true in our multi-agent agent.

\begin{verbatim}
    te1 = xferToGreeting >> sayHello + xferToWeather
\end{verbatim}
	
This first temporal expression states that if the current event is a transfer to the Greeting agent, then the next event should be a call to the \texttt{say\_hello} tool, and the following event should be a transfer back to the main Weather agent.  Essentially, it says that a transfer to the Greeting agent should produce the "hello" message and then control should transfer back to the main Weather agent.

\begin{verbatim}
    te2 = xferToFarewell >> sayGoodbye + xferToWeather
\end{verbatim}

This second temporal expression is similar to the first, except is states that if control is transferred to the Goodbye agent, we expect the "goodbye" message and then control should pass back to the main Weather agent.

\begin{verbatim}
    te3 = xferToWeather >> (getWeather | xferToGreeting | xferToFarewell)
\end{verbatim}

This last temporal expression states that if control is passed back to the main Weather agent, then the next event should either be a call to \texttt{get\_weather} or a transfer to one of the sub agents.


\section{Experimental Results}

A conversational script was developed that was used to drive the three-agent system.  In ten executions of the system with a strong LLM, the system behaved as expected.  With a smaller LLM substituted for the Greeting and Farewell agents, mistakes sometimes occurred.  The temporal expression assertions flagged these errors.  This section shows how.

The short driving script is shown below.

\begin{quote}
  \textbf{Hello. I would like a greeting, and then I want to know the weather.}\\
  \textbf{What is the weather in New York?}
\end{quote}

This short script presents a challenge to the system.  The first line contains both a greeting and a request, and the system needs to plan how to use the Greeting agent and then return.

The script was fed to the ADK Web UI.  This interactive UI shows in detail the LLM conversations, the tool calls and the input and output natural language phrases.  That information is presented here, as is possible in printed form.

\subsection{Using a Strong LLM}

In the first configuration, the model 'claude-3-7-sonnet-20250219' was used in all three agents.  This is a very powerful model and it was hosted in the cloud, at Anthropic.  The script above was applied to the system ten times.  The results were very coherent every time.

An illustration of the conversation including user input, agent output and tool call traces is shown below.  \textbf{User input} is in bold, \textit{agent output} in italics and \texttt{tool calls} in fixed width.

\noindent
\\
\textbf{Hello. I would like a greeting, and then I want to know the weather.}\\
\textit{I'll help you with both your greeting and weather request, but first I'll need to transfer your greeting request to our specialized greeting agent.}\\
\texttt{transfer\_to\_greeting}\\
\textit{I'd be happy to provide you with a greeting!}\\
\texttt{say\_hello}\\
\textit{Now that I've greeted you, I'll transfer you back to the weather agent who can help with your weather request.}\\
\texttt{transfer\_to\_weather}\\
\textit{Hello! Now that you've been greeted, I can help you with the weather information you requested. To provide an accurate weather report, I'll need to know which city you're interested in. Could you please let me know which location you'd like the weather report for?}\\
\\
\textbf{What is the weather in New York?}\\
\texttt{get\_weather}\\
\textit{The weather in New York is currently sunny with a temperature of 25°C. It's a beautiful day there!  Is there anything else you'd like to know about the weather in New York or any other city?}\\

When using this powerful LLM in each of the agents, each natural language input is entirely satisfied.  The first input, consisting of a greeting and a weather request, is decomposed into a transfer to the Greeting agent, a \texttt{say\_hello} tool call and then an immediate return to the main Weather agent.  Claude Sonnet also produces a very clear explanation of what it is doing at each step.

There are no assertion failures of the temporal expressions.   One illustrative match is reproduced below and shows that after control was passed to the Greeting agent, the \texttt{sayHello} tool call occurred, and then transfer back to the main Weather agent occurred.

\begin{verbatim}
MATCH
>> (1/3) (10/30) 0 {}
  isXferToGreeting (1/1) (10/10) 0 {}
  + (2/3) (20/30) 0 {}
    isSayHello (2/2) (20/20) 0 {}
    isXferToWeather (3/3) (30/30) 0 {}
\end{verbatim}

\subsection{Substituting Smaller LLMs}

In the second configuration, the Greeting and Farewell agents were configured to use "ollama\_chat/llama3.2:latest" running locally on a Macbook Pro.  This model is only 2.0 GB, as reported by Ollama.  The script was run against this system ten different times.  In this configuration, there were occasional failures.  This section illustrates the type of failures and how they were detected by temporal expressions.

\subsubsection{Successful Example}

In 7 out of 10 runs of the system with the less powerful LLMs, the system did correctly transfer to the Greeting agent and then transferred back to the Weather agent.  The following exchange is representative of the session.

\noindent
\\
\textbf{Hello. I would like a greeting, and then I want to know the weather.}\\
\textit{I'll need to delegate the greeting part of your request, and then I can help you with the weather information.}\\
\texttt{xfer\_to\_greeting}\\
\texttt{say\_hello}\\
\texttt{xfer\_to\_weather}\\
\textit{Hello, there!}\\
\textit{Now that I've taken care of the greeting, I'd be happy to help you with the weather. Could you please let me know which city you'd like the weather information for?}\\
\\
\textbf{What is the weather in New York?}\\
\texttt{get\_weather}\\
\textit{The weather in New York is sunny with a temperature of 25°C. Is there anything else you'd like to know about the weather in New York or any other city?}\\

This conversation is sensible, and the temporal expression assertions did not flag any problems.

\subsubsection{Failing Example}

In 3 out of 10 runs of the system, the weaker LLM could not complete the two tasks of calling the \texttt{say\_hello} tool and then transferring control back to the Weather agent.  The following exchange is representative of how this problem manifested itself.  The temporal expression assertions caught the problems and reported them.

\noindent\\
\textbf{Hello. I would like a greeting, and then I want to know the weather.}\\
\texttt{xfer\_to\_greeting}\\
\texttt{say\_hello}\\
\texttt{xfer\_to\_greeting}\\
\textit{I do not have the 'weather\_agent\_v2' agent to transfer this request to.
  However, I can answer your original question about the weather by saying:
  Unfortunately, I don't have real-time access to current weather conditions.
  If you'd like, I can suggest some ways for you to find out the current weather in your area, such as checking a weather website or app on your phone.}\\
\\
\textbf{What is the weather in New York?}\\
\texttt{xfer\_to\_weather}\\
\textit{I'll get the weather information for New York for you right away.}\\
\texttt{get\_weather}\\
\textit{The weather in New York is currently sunny with a temperature of 25°C. It's a beautiful day in the city!}\\

This conversation exhibited some problems.  The main agent decomposed the first request into two parts.  But after it transfered control to the Greeting agent (powered by llama3.2), that agent called the \texttt{say\_hello} tool but did not transfer control back to the Weather agent.  It looks like the Greeting agent attempted to transfer control, but transferred it to the wrong agent, the Greeting agent again.

The assertions monitoring the temporal assertions were clear about the failures.  There were two failure traces produced.  The first one starts at event 1 and ends at event 3 and states that the sequence

\begin{verbatim}
te1 = xferToGreeting >> sayHello + xferToWeather
\end{verbatim}
   
\noindent
was not obeyed.  That is in fact what is shown in the conversation.  While there was a transfer to the Greeting agent and a call to \texttt{say\_hello}, there was not a correct transfer back to the Weather agent.

\begin{verbatim}
FAILURE
>> (1/3) (10/30) 0 {}
  isXferToGreeting (1/1) (10/10) 1 {}
  + (2/3) (20/30) 0 {}
    isSayHello (2/2) (20/20) 1 {}
    isXferToWeather (3/3) (30/30) 0 {}
Temporal expression always assertion failed.
\end{verbatim}

The second failure starts at event 3 and ends at event 4.  It was also checking the sequence that begins with \texttt{xfer\_to\_greeting}.  In this case, it catches the fact that after the second transfer to the Greeting agent, there was no tool call to \texttt{say\_hello}.

\begin{verbatim}
FAILURE
>> (3/4) (30/40) 0 {}
  isXferToGreeting (3/3) (30/30) 1 {}
  + (4/4) (40/40) 0 {}
    isSayHello (4/4) (40/40) 0 {}
Temporal expression always assertion failed.
\end{verbatim}

\section{Conclusion}

As AI agents increasingly rely on natural language for both input processing and application logic, they exhibit behaviors that can be difficult to predict or control. Traditional software testing approaches fall short when dealing with the dynamic, language-driven nature of these systems.

This paper presents a runtime monitoring approach using temporal expressions to specify and verify the behavioral constraints of AI agentic systems. The Oroboro temporal expression package operates alongside agents, capturing event traces and evaluating them against predefined assertions. When violations occur, the system can halt execution or provide feedback for agent recovery planning.

The evaluation demonstrates that temporal expressions effectively capture sequencing errors and behavioral anomalies without requiring natural language understanding.  This event-driven monitoring proves particularly valuable for AI agents because it focuses on observable behaviors rather than attempting to interpret the semantic content of natural language interactions.

However, important limitations remain. While temporal expressions handle sequencing and workflow violations well, they cannot address errors requiring semantic analysis of text, images, or conversation context. Additionally, the current approach requires developers to manually instrument events and craft assertions, potentially creating a significant development burden.

Future agent development frameworks must treat safety assertions as first-class components, not afterthoughts. Advanced tooling should automatically generate event predicates from common agent patterns, provide built-in methods for accessing textual and media content within assertions, and offer libraries of reusable temporal patterns for common agent behaviors. Such frameworks would make creating comprehensive safety checks as straightforward as writing the core agent logic itself.

The path toward trustworthy AI agents requires multiple complementary approaches. Temporal expression monitoring addresses a crucial piece of this puzzle by providing runtime behavioral verification, but it must be combined with semantic analysis tools and robust development practices to ensure truly safe agent deployments.

\printbibliography

\end{document}